\documentclass[letterpaper,10pt,conference]{ieeeconf}

\IEEEoverridecommandlockouts
\usepackage{amsmath}
\usepackage{amssymb}
\usepackage{algorithm}
\usepackage{algorithmic}
\usepackage{booktabs}
\usepackage{threeparttable}
\usepackage{capt-of}
\usepackage{graphicx}
\usepackage[table]{xcolor}
\makeatletter
\let\NAT@parse\undefined
\makeatother
\usepackage[hidelinks]{hyperref}
\hypersetup{
  pdftitle={DexMimic: A Generalist Dexterous Hand-Object Tracking Policy via Value-Guided Reset Allocation},
  pdfauthor={Zhexi Luo, Zhixuan Xu, Rui Yang, Zeyu Xiao, Yiwen Hou, Lin Shao}
}

\makeatletter
\long\def\@makecaption#1#2{%
  \par\addvspace{6pt}
  \setbox\@tempboxa\hbox{\footnotesize\normalfont #1: #2}%
  \ifdim\wd\@tempboxa>\hsize
    {\footnotesize\normalfont\raggedright\noindent #1: #2\par}%
  \else
    \hb@xt@\hsize{\hfil\box\@tempboxa\hfil}%
  \fi}
\makeatother

\definecolor{dexcolor}{HTML}{386CB0}
\definecolor{mimiccolor}{HTML}{D56B35}
\definecolor{bestcell}{HTML}{DFF0D8}
\definecolor{weakcell}{HTML}{F8D7DA}
\newcommand{\dexmimic}{\textcolor{dexcolor}{Dex}\textcolor{mimiccolor}{Mimic}}

\title{\textbf{\dexmimic}: A Generalist Dexterous Hand--Object Tracking Policy\\
via Value-Guided Reset Allocation}

\author{Zhexi Luo$^{1,2}$, Zhixuan Xu$^{1,2}$, Rui Yang$^{1,2}$,
Zeyu Xiao$^{1,2}$, Yiwen Hou$^{1,2}$, Lin Shao$^{1,2}$\\[3pt]
{\small $^{1}$National University of Singapore (NUS)\quad $^{2}$RoboScience}\\[3pt]
{\small \url{https://zhexiluo.github.io/DexMimic/}}}

\IEEEaftertitletext{%
  \begin{minipage}{\textwidth}
    \centering
    \includegraphics[width=\textwidth]{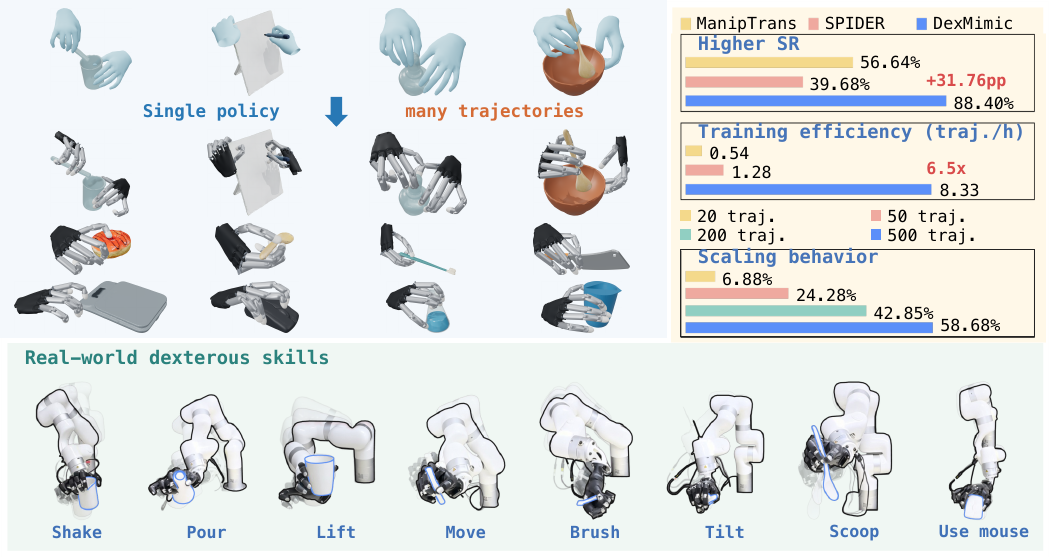}
    \captionof{figure}{DexMimic learns a generalist dexterous hand--object tracking policy from human demonstrations through Value-Guided Reset Allocation. It consistently outperforms trajectory-specific baselines in tracking success and training efficiency, with zero-shot generalization improving as co-training scales. Real-world experiments validate DexMimic's ability to perform diverse dexterous manipulation skills.}
    \label{fig:teaser}
  \end{minipage}\vspace{6pt}%
}

\begin{document}
\bstctlcite{compactctl}
\raggedbottom
\brokenpenalty=10000
\setlength{\textfloatsep}{14pt plus 2pt minus 2pt}
\setlength{\dbltextfloatsep}{14pt plus 2pt minus 2pt}

\maketitle
\thispagestyle{plain}
\pagestyle{plain}

\begin{abstract}
Achieving human-like dexterous manipulation remains a central challenge in robotics. Learning from human hand--object interaction (HOI) datasets offers a promising pathway to diverse dexterous skills, yet existing physics-based approaches train specialist policies for individual trajectories, thereby limiting skill reuse at scale. We introduce \textbf{\dexmimic}, a generalist dexterous hand--object tracking policy jointly trained across diverse HOI trajectories. Efficient joint training requires allocating each rollout to both an undertrained phase and a starting state that supports continued tracking. DexMimic realizes this insight through \textbf{Value-Guided Reset Allocation}, which adapts the initialization distribution online to prioritize difficult phases and select reset states with higher expected tracking returns. We further introduce a benchmark spanning training, fine-tuning, and generalization. Experiments show improved tracking success and training efficiency, faster fine-tuning, and the potential for zero-shot generalization to unseen trajectories and objects; larger training sets further improve efficiency and generalization. Finally, we validate DexMimic on a real-world dexterous manipulation platform.
\end{abstract}

\section{Introduction}
Recent advances in physics-based humanoid control show that a single reference-conditioned policy can learn diverse motor skills from large-scale motion datasets and reuse the learned control prior beyond individual training motions~\cite{peng2018deepmimic,luo2023phc,liao2025beyondmimic,luo2024pulse,tessler2024maskedmimic,he2025omnih2o,ma2026robusttracking}. Extending this paradigm to dexterous hand--object interaction could enable robots to acquire diverse manipulation skills from human demonstrations~\cite{shaw2023videodex,wang2024dexcap,mu2026deximit,tao2025dexwild}. These demonstrations capture fine-grained finger coordination, contact transitions~\cite{fan2023arctic,liu2024taco,taheri2020grab}, and regrasping behaviors that are difficult to specify through task-specific rewards. A generalist hand--object tracking policy could provide a reusable low-level control interface conditioned on human demonstrations. Realizing this vision requires bridging the substantial embodiment gap between human and robot hands in morphology, actuation, and contact dynamics~\cite{handa2020dexpilot,li2025maniptrans,mandi2025dexmachina,wei2025omnidexgrasp,liang2026spatialbrain,zhao2024dexh2r,xin2025retargeting,qin2023anyteleop,sivakumar2022telekinesis}.

Recent physics-based reinforcement learning methods have made substantial progress in bridging this gap for individual demonstrations, using general tracking objectives for hand motion, contact states, and object motion~\cite{li2025maniptrans,mandi2025dexmachina}. However, these methods typically train a specialist policy for each trajectory or a small trajectory subset. This specialist formulation limits the reuse of learned control priors across trajectories and incurs repeated training costs as the demonstration set grows. To overcome these limitations, we investigate what prevents existing trajectory-specific training pipelines from scaling to a generalist policy that jointly learns diverse hand--object interaction (HOI) trajectories and transfers the learned interaction prior to unseen references.

We find that Reference State Initialization (RSI)~\cite{peng2018deepmimic}, a standard mechanism for initializing training episodes, becomes a key bottleneck when diverse trajectories share a limited rollout budget. RSI initializes each episode with a state sampled from a human motion retargeted to the robot's morphology. This initialization scheme introduces inefficiencies in both the choice of starting states and the allocation of training across reference phases. First, kinematically retargeted states may contain physically inconsistent contacts, causing \emph{fast-fail resets} before useful interaction experience is collected. Second, uniform timestamp sampling oversamples simple phases while undertraining difficult phases. As the trajectory set grows, these inefficiencies reduce the useful interaction experience available to each trajectory, limiting the scalability of joint policy learning.

We introduce \textbf{\dexmimic}, a generalist dexterous hand--object tracking policy built on the insight that effective joint training requires adapting the initialization distribution alongside policy learning. Its \textbf{Value-Guided Reset Allocation} uses critic estimates to prioritize under-learned reference phases and select starting states that support continued tracking from phase-conditioned reset buffers. Each training iteration collects rollouts from the current initialization distribution and uses the same experience to update the policy and critic, refine reset candidates, and adjust phase priorities, thereby shaping the distribution for subsequent rollouts. On 50 challenging bimanual tasks, DexMimic achieves 88.40\% success versus 56.64\% for ManipTrans, while reducing average training time per trajectory by a factor of 15.5. Scaling co-training from 20 to 500 trajectories raises zero-shot success on unseen objects from 2.00\% to 51.65\%, demonstrating an increasingly transferable interaction prior. We further validate DexMimic on a real-world robotic platform.

In summary, our contributions are threefold: \textbf{(1)} We introduce \textbf{\dexmimic}, a generalist dexterous hand--object tracking policy with \textbf{Value-Guided Reset Allocation}, which adapts reset state and phase allocation alongside policy learning. \textbf{(2)} We introduce a benchmark for generalist hand--object tracking that evaluates joint-training performance and efficiency, fine-tuning on unseen references, and zero-shot generalization to unseen trajectories and objects. \textbf{(3)} We demonstrate improved tracking success and training efficiency, faster adaptation, and stronger zero-shot generalization as training scales. We further validate DexMimic on a real-world robotic platform.

\section{Related Work}
\subsection{Reference tracking in physics simulation}
Human hand--object demonstrations provide rich references for learning contact-rich manipulation in physics simulation~\cite{qin2021dexmv,mandikal2022dexvip,zhou2026bidexhd,singh2025hop,chen2024objectcentric}. ManipTrans~\cite{li2025maniptrans} uses reinforcement learning to transfer bimanual skills through residual control, while DexMachina learns functional retargeting through a curriculum over virtual object controllers~\cite{mandi2025dexmachina}. Further advances include adaptive trade-offs between object tracking and motion fidelity~\cite{liang2026contrack}, reference-scoped exploration~\cite{xu2025dexplore}, and generalizable tracking~\cite{liu2025dextrack}. Physics-based trajectory optimization also enables the conversion of human demonstrations into dynamically feasible robot motions~\cite{pan2025spider}.

In humanoid control, shared policies support diverse whole-body skills~\cite{ze2025twist,peng2022ase,juravsky2024superpadl,wu2025uniphys}, and SONIC~\cite{luo2025sonic} demonstrates the benefits of scaling data, model capacity, and computation for motion tracking. These advances motivate learning reusable control capabilities from diverse references. DexMimic takes a step toward this goal for dexterous hand--object interaction through efficient joint training across diverse HOI trajectories.

\subsection{Reset distributions for reinforcement learning}
The initialization distribution shapes the experience available to a policy and thereby influences exploration and learning efficiency~\cite{tavakoli2019restart,ecoffet2021goexplore,tao2024rfcl,nair2017overcoming}. Curricula over starting states and demonstration-guided resets have shown the benefits of organizing where learning begins~\cite{florensa2017reverse,resnick2018backplay,uchendu2023jumpstart,popov2017dataefficient,bauza2024demostart,dai2021aced,hermann2020acgd}. In physics-based tracking, reference initialization, reuse of simulated states, and adaptive sampling further improve access to useful training experience~\cite{peng2018deepmimic,xu2025intermimic,xu2025dexplore,liao2025beyondmimic,liang2026contrack,xu2026demobot}. DexMimic treats initialization as an evolving component of learning: a continually updated critic guides the distribution toward under-learned phases and starting states that support continued tracking, allowing reset allocation to adapt alongside policy learning.

\section{Method}
\label{sec:method}

\begin{figure*}[t]
  \centering
  \includegraphics[width=\textwidth]{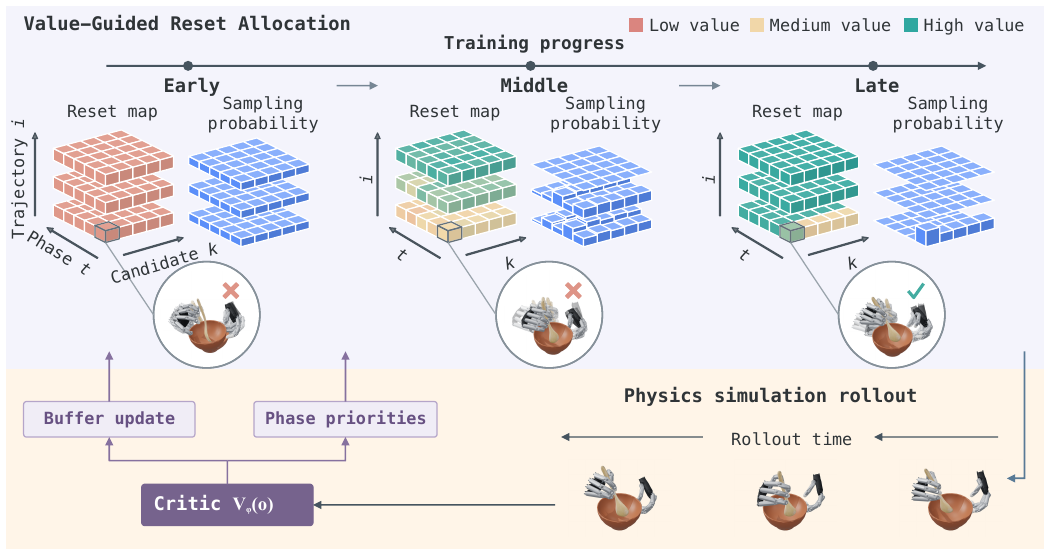}
  \caption{Overview of Value-Guided Reset Allocation in DexMimic. The reset map stores candidate states and their critic values for each trajectory and reference phase. As training progresses, reset phase allocation prioritizes under-learned phases, while reset state allocation favors candidates with higher expected tracking returns within each selected phase. Physics rollouts from the sampled initializations update the policy and critic, replenish the reset map, and refine phase sampling probabilities, forming an adaptive training loop.}
  \label{fig:method}
\end{figure*}

DexMimic learns a shared reference-conditioned tracking policy across
diverse HOI trajectories in physics simulation. Our key insight is that
achieving effective policy learning at scale hinges on optimizing the
initialization distribution. To this end, Value-Guided Reset Allocation
uses critic estimates to adapt the initialization distribution alongside
policy learning. Reset state allocation selects starting states that support
continued tracking, while reset phase allocation directs training toward
under-learned reference phases. Fig.~\ref{fig:method} illustrates the
resulting training loop.

\subsection{Problem Formulation}

Given a set of reference trajectories $\mathcal D=\{\tau_i\}_{i=1}^{N}$
from HOI datasets, our goal is to learn a shared tracking policy.
Each trajectory $\tau_i=\{\hat{x}_{i,t}\}_{t=0}^{T_i}$ specifies the
target hand--object state and contact geometry at each reference frame.
We formulate this task as a reference-conditioned Markov decision
process (MDP),
$\mathcal M=\langle\mathcal S,\mathcal A,\mathcal P,\mathcal R,
\rho_0,\gamma\rangle$.
The task state $(i,t,s_t)\in\mathcal S$ includes the reference trajectory
index $i$, reference frame $t$, and hand--object simulator state $s_t$.
$\mathcal A$, $\mathcal P$, $\mathcal R$,
$\rho_0$, and $\gamma$ denote the continuous action space, simulation
dynamics, reward function, initialization distribution, and discount
factor, respectively.

At the start of each episode, we sample a reference trajectory $i$,
a starting frame $t_0$, and an initial simulator state $s_{t_0}$ jointly
from $\rho_0$. We optimize the shared policy $\pi_\theta(a_t\mid o_t)$
to maximize the expected discounted tracking return:
\begin{equation}
J(\theta;\rho_0)
=
\mathbb E_{\substack{
(i,t_0,s_{t_0})\sim\rho_0\\
\pi_\theta
}}
\left[
\sum_{h=0}^{H-1}\gamma^h r_{t_0+h}
\right].
\label{eq:method-1}
\end{equation}
Here, $o_t$ is the reference-conditioned observation, $r_t$ is the tracking
reward, and $H$ is the episode length until reference completion or
early termination.

\subsection{Reference-Conditioned Tracking Policy}

\noindent\textbf{Observation.}
The policy observation $o_t$ includes recent hand--object states, object
geometry and mass properties, and future reference targets. The state
history includes wrist position and orientation, finger joint angles,
object position and orientation, their corresponding velocities, fingertip
contact forces, and object translation relative to the wrist. Future
reference inputs describe target hand--object states, tracking errors,
and desired contact patterns. The observation also includes the kinematic
base action and the remaining reference steps $\ell_t$.

\noindent\textbf{Action.}
The policy outputs continuous residual actions for wrist translation,
wrist rotation, and finger joint positions. These residuals are combined
with a kinematic base action derived from retargeted hand references to
produce wrist forces and torques and finger position targets. Object
motion follows contact dynamics. Detailed observation features and action
parameterization are provided in the supplementary material.

\noindent\textbf{Reward.}
We use a task-agnostic tracking reward shared across all references,
\begin{equation}
\begin{aligned}
r_t={}&
\lambda_h r_t^{\mathrm{hand}}
+\lambda_o r_t^{\mathrm{object}}
+\lambda_v r_t^{\mathrm{velocity}}\\
&+\lambda_c r_t^{\mathrm{contact}}
+\lambda_e r_t^{\mathrm{energy}},
\end{aligned}
\label{eq:method-4}
\end{equation}
covering hand and object pose tracking, velocity matching,
reference-guided contact, and mechanical-power regularization.
Episodes terminate upon reference completion or excessive tracking error.

\noindent\textbf{Network and training.}
The actor $\pi_\theta$ and critic $V_\phi$ share a multilayer perceptron (MLP) backbone and
are jointly trained using standard Proximal Policy Optimization (PPO)~\cite{schulman2017ppo}.
Beyond policy learning, the critic provides value estimates for adapting
the initialization distribution.

\subsection{Value-Guided Reset Allocation}
\label{sec:reset-allocation}

At scale, fast-fail resets and repeated sampling of well-learned phases
consume a limited rollout budget, leaving insufficient experience for
difficult interactions. We address this by adapting two initialization
decisions: which reference phase to practice and which physical state to
start from. At training iteration $m$, we factorize the initialization
distribution as
\begin{equation}
\rho_{0,m}(i,t,s)
=
p_{\mathrm{phase},m}(i,t)\,
p_{\mathrm{state},m}(s\mid i,t).
\label{eq:method-5}
\end{equation}
Here, $p_{\mathrm{phase},m}$ selects a reference phase $(i,t)$, and
$p_{\mathrm{state},m}$ selects a starting state conditioned on that phase.
By estimating the expected tracking return, the critic identifies
promising starting states and under-learned phases. As learning progresses,
rollout experience updates the policy and critic while refining reset
candidates and phase priorities. The initialization distribution thus
adapts alongside the policy and critic, continually directing training
toward difficult interactions from states that support continued tracking.
We omit the iteration subscript below for clarity.

\begin{algorithm}[t]
\caption{DexMimic training pipeline}
\label{alg:training}
\begin{algorithmic}[1]
\STATE \textbf{Input:} HOI reference trajectories and training iterations
\STATE \textbf{Initialize:} Policy, critic, and reset buffers
\FOR{each training iteration}
    \STATE \textit{// Value-Guided Reset Allocation}
    \STATE Sample $(i,t)\sim p_{\mathrm{phase}}$
    \STATE Sample $s\sim p_{\mathrm{state}}(\cdot\mid i,t)$ using Eq.~\eqref{eq:method-7}
    \STATE \textit{// Physics-based policy learning}
    \STATE Collect physics rollouts from the sampled reset
    \STATE Update the policy and critic with PPO
    \STATE \textit{// Online reset adaptation}
    \STATE Update reset buffers and phase priorities from rollouts
    \STATE Form the next initialization distribution
\ENDFOR
\STATE \textbf{Output:} Trained generalist tracking policy
\end{algorithmic}
\end{algorithm}

\subsubsection{Reset state allocation}

Kinematically retargeted initializations can contain physically inconsistent
contacts, causing fast-fail resets before useful interaction experience is
collected. We therefore maintain a reset buffer for each reference phase
$(i,t)$ to reuse promising starting states discovered during physical rollouts:
\begin{equation}
\mathcal B_{i,t}
=
\left\{
\left(s_{i,t}^{k},v_{i,t}^{k}\right)
\right\}_{k=1}^{K}.
\label{eq:method-6}
\end{equation}
Collectively, these phase-conditioned buffers form a reset map over
the training trajectories.
Each buffer holds $K$ candidate states and their critic scores; each state
stores the hand--object configurations, velocities, and controller targets
needed to restore the simulation. We initialize every slot with the
phase-aligned reference state $s_{i,t}^{\mathrm{ref}}$ and its critic score.

Given a reference phase $(i,t)$, we sample $s_{\mathrm{reset}}\sim
p_{\mathrm{state}}(\cdot\mid i,t)$ using the nonnegative candidate weights
$w_k=\max(v_{i,t}^{k},0)$:
\begin{equation}
s_{\mathrm{reset}}
=
\begin{cases}
s_{i,t}^{\mathrm{ref}},
& \text{with probability }\alpha,
\\[5pt]
s_{i,t}^{k},
& \text{with probability }(1-\alpha)
\dfrac{w_k}{\sum_{j=1}^{K}w_j}.
\end{cases}
\label{eq:method-7}
\end{equation}
Critic-weighted sampling favors candidates with higher expected tracking
returns within the selected phase. With probability $\alpha$, we instead
sample the fixed reference anchor $s_{i,t}^{\mathrm{ref}}$, helping limit
drift away from the reference demonstrations. The subsequent tracking targets
remain those of the original reference trajectory.

As training progresses, states collected during physical rollouts are
evaluated as reset candidates for the corresponding reference phases.
For a candidate $s$, we record
its critic score $v=V_\phi(o)$ and evaluate its subsequent tracking
duration. To retain candidates that support continued tracking, we require
their survival fraction $c(s)$---the fraction of remaining reference steps
tracked from $s$ before rollout termination---to exceed $c_{\min}$. An
eligible candidate replaces
the lowest-scored slot in $\mathcal B_{i,t}$ only when its score exceeds
the stored score. During buffer maintenance, stored scores decay as
$v\leftarrow(1-\delta)v$, allowing candidates generated by later
policies to enter the buffer. Reference anchors remain unchanged.
Together, candidate replacement and score decay adapt $p_{\mathrm{state}}$
throughout training, extending reference-based initialization with reusable
rollout experience.

\subsubsection{Reset phase allocation}

To avoid repeatedly sampling well-learned phases, reset phase allocation
uses critic estimates to prioritize under-learned interactions. We partition
the reference trajectories into temporal bins, indexed globally by
$b=1,\ldots,B$. To account for differences in remaining reference length,
we normalize the critic value at each rollout step by the sum of the
remaining discount weights:
\begin{equation}
\nu_t
=
\frac{V_\phi(o_t)}
{\sum_{h=0}^{\ell_t-1}\gamma^h},
\label{eq:method-8}
\end{equation}
where $\ell_t>0$ is the number of remaining reference steps.
Averaging normalized values over rollout steps
assigned to bin $b$ yields $\bar\nu_b$, a critic-based estimate of the
policy's tracking performance in that bin.
We compute a difficulty score $f_b$ by subtracting $\bar\nu_b$ from
the theoretical maximum normalized value determined by the reward
definition. Larger scores indicate phases that remain under-learned.

Poor tracking in a difficult phase may also result from insufficient
preparation in preceding phases. We therefore propagate each bin's
difficulty score backward over a window of $n$ bins within the same
reference trajectory, with geometrically decaying weights:
\begin{equation}
\tilde f_b
=
\sum_{d=0}^{n-1}\lambda^d f_{b+d},
\quad n=5,\quad 0<\lambda<1,
\label{eq:method-9}
\end{equation}
Here, $b+d$ denotes the current or a later bin within the same reference
trajectory, and terms crossing trajectory boundaries are omitted.
This raises reset priorities before difficult interactions, giving the
policy opportunities to learn the preparation needed to enter them.

To reduce abrupt changes in reset priorities, we smooth the propagated
difficulty scores across training iterations using an exponential moving
average (EMA):
\begin{equation}
\bar f_b
\leftarrow
(1-\beta)\bar f_b+\beta\tilde f_b,
\qquad 0<\beta<1,
\label{eq:method-10}
\end{equation}
where $\beta$ weights the newly propagated score relative to the previous
estimate. We normalize the smoothed scores across all bins to obtain
\begin{equation}
p_{\mathrm{bin}}(b)
=
\frac{\bar f_b}{\sum_{j=1}^{B}\bar f_j}.
\label{eq:method-11}
\end{equation}
We sample a bin from $p_{\mathrm{bin}}$, then uniformly sample a reference
frame $(i,t)$ within it. Thus,
$p_{\mathrm{phase}}(i,t)=p_{\mathrm{bin}}(b(i,t))/n_{b(i,t)}$,
where $b(i,t)$ is the bin containing $(i,t)$ and $n_b$ is its number
of reference frames. We then sample the reset state from
$p_{\mathrm{state}}(\cdot\mid i,t)$ in Eq.~\eqref{eq:method-7},
including its reference-anchor branch.
As new rollout values are collected, phase priorities adapt to the policy's
evolving tracking performance.

\subsubsection{Online training loop}

Algorithm~\ref{alg:training} summarizes how policy learning and reset
allocation are coupled through shared rollout experience. At each iteration,
DexMimic samples reference phases and starting states from the current
initialization distribution, then collects physics rollouts to update the
policy and critic with PPO. The same rollouts refine reset candidates and
phase priorities, directing subsequent training toward under-learned
interactions from states that support continued tracking.

\begin{figure*}[t]
  \centering
  \includegraphics[width=\textwidth]{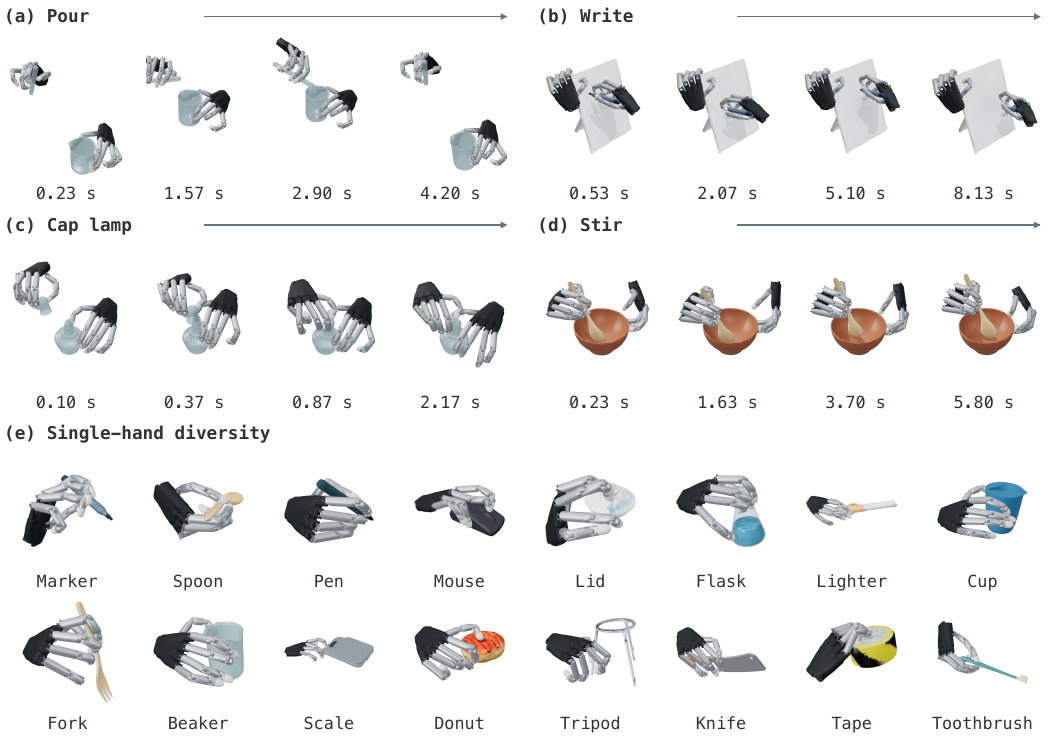}
  \caption{Qualitative simulation results of DexMimic. The top two rows show four bimanual
  rollouts, with four temporally ordered frames per rollout from left to right.
  The bottom two rows show representative states from 16 single-hand trajectories.}
  \label{fig:simulation-rollouts}
\end{figure*}

\section{Experiments}
\subsection{Experimental Settings}
\label{sec:experimental-settings}

\noindent\textbf{Benchmark.}
We comprehensively evaluate DexMimic on 1,944 dexterous manipulation tasks
involving 140 objects from the OakInk-V2~\cite{zhan2024oakink2} and HOT3D~\cite{banerjee2025hot3d} human hand--object
interaction (HOI) datasets,
comprising 1,302 single-hand trajectories and 642
bimanual trajectories involving rigid objects. We construct reference sets for
training, fine-tuning, and zero-shot evaluation, with the zero-shot evaluation
set organized into three levels of increasing difficulty. All main benchmark
experiments use the Wuji dexterous hand, while results on additional hand
embodiments are reported separately.

\noindent\textbf{Training and evaluation protocol.}
We train policies using PPO in Isaac Gym~\cite{makoviychuk2021isaacgym} with 65,536 parallel environments
across eight NVIDIA A100 GPUs. Value-Guided Reset Allocation is applied only during
training to determine the reset phases and initialization states. During
evaluation, we use Reference State Initialization (RSI) to initialize the
hand and object at the first frame of the original reference and evaluate
each task with 50 complete rollouts.

\begingroup
\displaywidowpenalty=10000
\setlength{\jot}{0pt}
\noindent\textbf{Metrics.}
We evaluate tracking performance using object rotation error, object
translation error, finger keypoint error, fingertip error, and success
rate (SR)~\cite{li2025maniptrans}. We additionally report the average
training time per trajectory. Let $\mathbf T_t=(\mathbf R_t,\mathbf p_t)$
denote the executed object pose and $\mathbf x_j^t$ the position of finger
keypoint $j$; barred quantities denote their references. With $J$ finger
keypoints and fingertip indices $\mathcal F$, the framewise errors are
\begingroup
\small
\begin{flalign}
&e_{\mathrm{rot}}^t
=\arccos\!\tfrac{\operatorname{tr}(\bar{\mathbf R}_t^\top\mathbf R_t)-1}{2},
\quad e_{\mathrm{trans}}^t
=\|\mathbf p_t-\bar{\mathbf p}_t\|_2,
&&\label{eq:tracking-errors}\\
&e_{\mathrm{finger}}^t
=\tfrac{1}{J}\textstyle\sum_{j=1}^{J}
\|\mathbf x_j^t-\bar{\mathbf x}_j^t\|_2,
\enspace e_{\mathrm{tip}}^t
=\tfrac{1}{|\mathcal F|}\textstyle\sum_{j\in\mathcal F}
\|\mathbf x_j^t-\bar{\mathbf x}_j^t\|_2.
&&\notag
\end{flalign}
\endgroup
\endgroup
Rotation error is reported in radians and position errors in millimeters.
A rollout succeeds if it completes the reference and its time-averaged
rotation, translation, finger keypoint, and fingertip errors are below
$30^\circ$, $3\,\mathrm{cm}$, $8\,\mathrm{cm}$, and $6\,\mathrm{cm}$,
respectively. In bimanual tasks, both hands must satisfy the success
criteria. SR is the fraction of successful rollouts. We report tracking
errors averaged over time and then over successful rollouts, with hand
errors averaged across hands in bimanual tasks.

\begin{figure*}[t]
  \centering
  \begin{minipage}[t]{0.7469\textwidth}
    \vspace{0pt}
  \includegraphics[width=\linewidth]{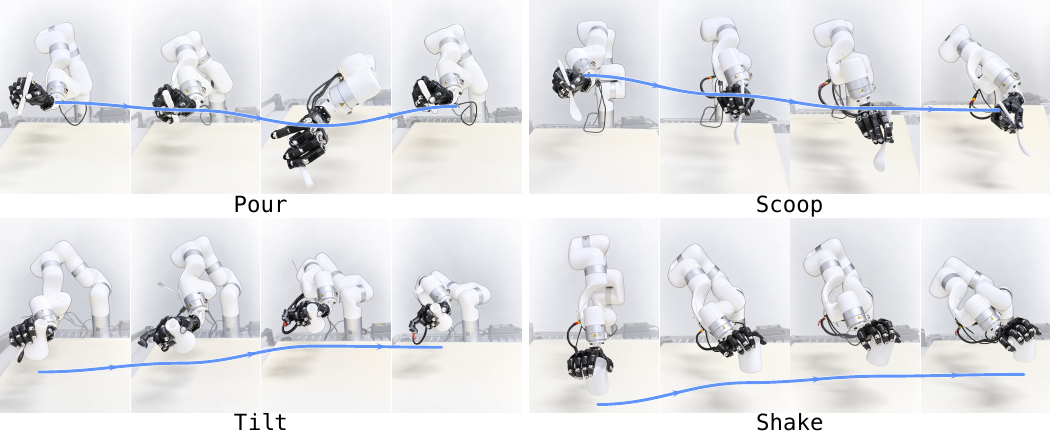}
  \caption{Real-world execution of four manipulation motions with DexMimic: pouring, scooping, tilting, and shaking. Frames progress from left to right within each sequence.}
  \label{fig:realworld-rollouts}
  \end{minipage}\hfill
  \begin{minipage}[t]{0.2331\textwidth}
    \vspace{0pt}
    \includegraphics[width=\linewidth]{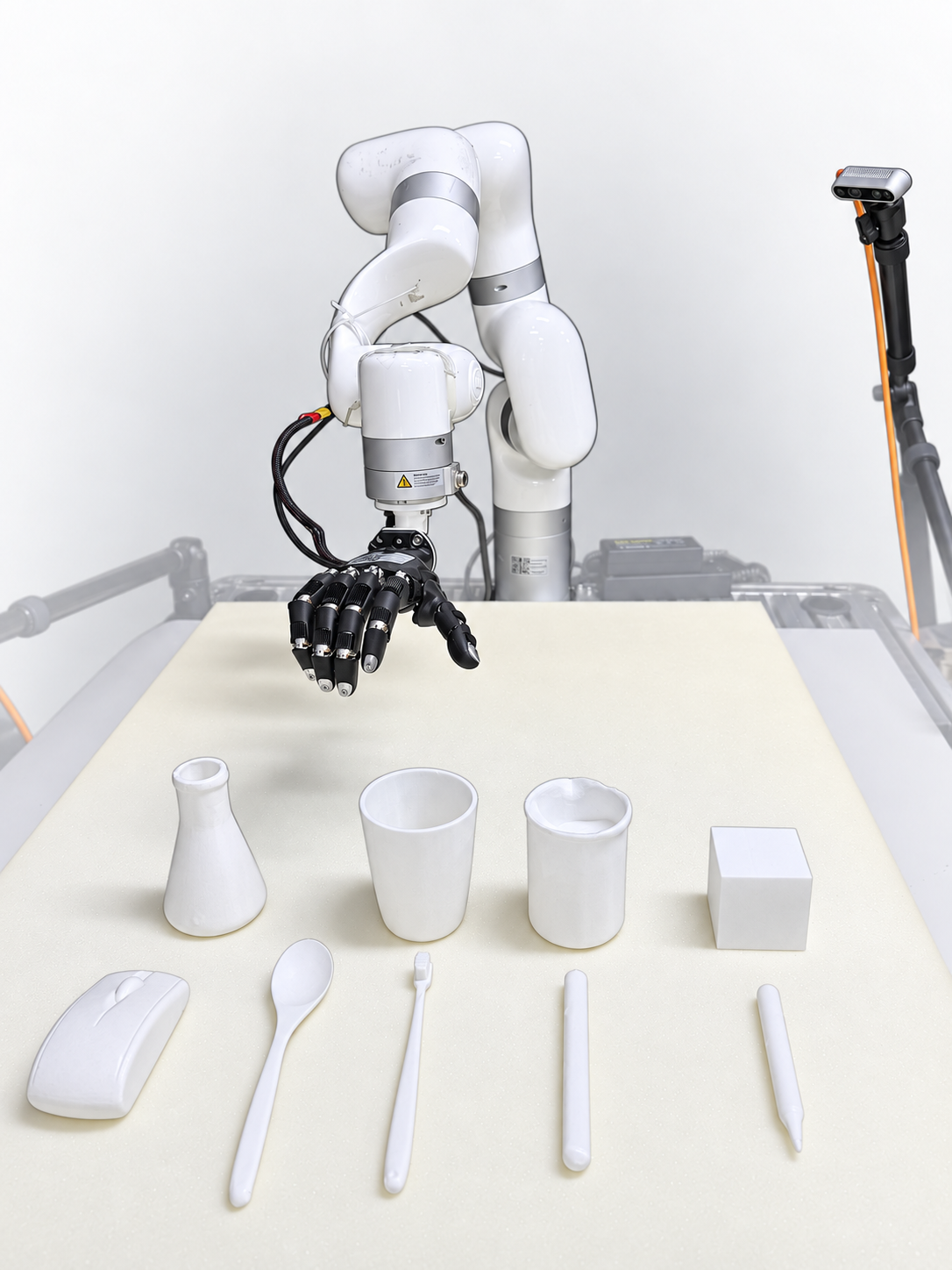}
    \caption{Real-world platform with eight 3D-printed OakInk objects.}
    \label{fig:realworld_setup}
  \end{minipage}
\end{figure*}

\subsection{Does DexMimic Improve Tracking Performance and Training
Efficiency over Trajectory-Specific Methods?}

We evaluate DexMimic on 50 challenging bimanual manipulation tasks from
OakInk-V2 and HOT3D, comparing against ManipTrans~\cite{li2025maniptrans} and SPIDER~\cite{pan2025spider} as representative
reinforcement learning and optimization baselines, respectively. ManipTrans
learns a trajectory-specific residual policy on top of a pretrained hand-motion
imitator to reproduce hand--object interactions. SPIDER directly optimizes a
dynamically feasible control sequence for each reference trajectory through
contact-guided sampling in parallel physics simulation. Both baselines train
or optimize separately for each trajectory, while DexMimic jointly trains a
single policy across all 50 trajectories. For a fair efficiency comparison,
we report the average training or optimization time per trajectory, dividing
DexMimic's total joint-training time by 50. All methods use identical initial
states, rollout counts, and success criteria during evaluation.

\begin{table}[t]
  \centering
  \begin{threeparttable}
    \centering
    \footnotesize
    \setlength{\tabcolsep}{5.5pt}
    \renewcommand{\arraystretch}{1.12}
    \begin{tabular}{@{}lcccccc@{}}
      Methods &
      \shortstack{SR $\uparrow$\\(\%)} &
      \shortstack{Time $\downarrow$\\(h)} &
      \shortstack{$E_r \downarrow$\\(rad)} &
      \shortstack{$E_t \downarrow$\\(mm)} &
      \shortstack{$E_j \downarrow$\\(mm)} &
      \shortstack{$E_{ft} \downarrow$\\(mm)} \\
      \midrule
      ManipTrans & 56.64 & 1.86 & 0.12 & 6.13 & 19.39 & \cellcolor{bestcell}14.42 \\
      SPIDER     & 39.68 & 0.78 & 0.35 & 12.14 & 38.97 & 27.48 \\
      \midrule
      DexMimic   & \cellcolor{bestcell}88.40 & \cellcolor{bestcell}0.12 & \cellcolor{bestcell}0.07 & \cellcolor{bestcell}3.76 & \cellcolor{bestcell}17.25 & 14.52 \\
    \end{tabular}
    \caption{Tracking performance and average training or optimization time
    per trajectory on 50 challenging bimanual tasks. $E_r$, $E_t$, $E_j$,
    and $E_{ft}$ denote object rotation, object translation, finger keypoint,
    and fingertip errors, respectively. Best values are highlighted in green.}
    \label{tab:main_results}
  \end{threeparttable}
\end{table}

Table~\ref{tab:main_results} summarizes the results on these 50 challenging
bimanual tasks. DexMimic achieves an 88.40\% success rate, outperforming
ManipTrans and SPIDER by 31.76 and 48.72 percentage points, respectively.
Its average training time is only 0.12~h per trajectory, compared with
1.86~h for ManipTrans and 0.78~h for SPIDER, corresponding to $15.5\times$
and $6.5\times$ lower time costs. DexMimic also achieves the lowest object
rotation and translation errors, at 0.07~rad and 3.76~mm, respectively.
These results demonstrate that a single shared tracking policy can efficiently
learn diverse dexterous manipulation tasks while achieving higher execution
success and more accurate object tracking than trajectory-specific methods.

Fig.~\ref{fig:simulation-rollouts} further illustrates DexMimic's tracking
behavior on representative single-hand and bimanual manipulation tasks.

\begin{figure*}[t]
  \centering
  \includegraphics[width=0.49\textwidth]{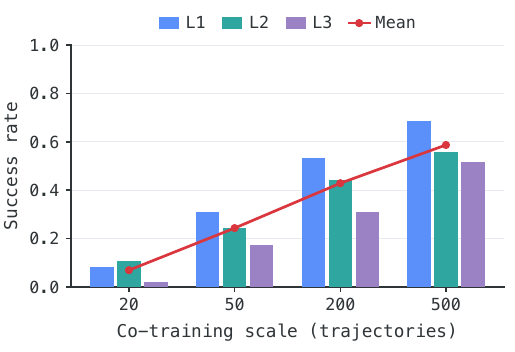}\hfill
  \includegraphics[width=0.49\textwidth]{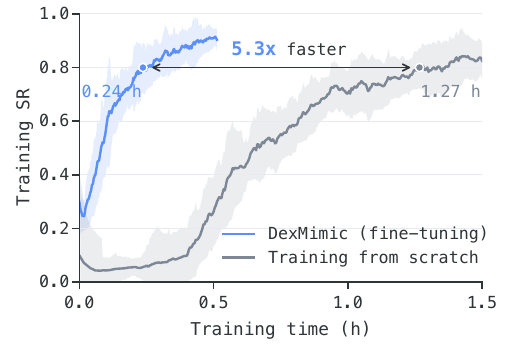}
  \par\vspace{1pt}
  \makebox[0.49\textwidth]{\footnotesize (a)}\hfill
  \makebox[0.49\textwidth]{\footnotesize (b)}
  \caption{Benchmark for generalization and adaptation.
    (a) Grouped bars show zero-shot SR at three generalization levels (L1--L3).
    The red line shows that the mean SR across these levels increases with
    co-training scale. (b) Fine-tuning the pretrained DexMimic base policy accelerates
    joint adaptation to 20 unseen trajectories.}
  \label{fig:generalization}
\end{figure*}

\subsection{Does DexMimic Learn a Transferable Interaction Prior?}

We introduce a benchmark to evaluate whether generalist co-training yields
a reusable interaction prior that supports both zero-shot generalization
and rapid adaptation, and how its transferability scales with training
data. The benchmark combines three levels of zero-shot evaluation with
fine-tuning on unseen references, covering transfer to new motions and
objects as well as the efficiency of acquiring new tracking behaviors.
For zero-shot evaluation, we train policies on nested subsets of 20, 50,
200, and 500 trajectories and freeze their parameters at test time.
We evaluate three levels: L1, similar trajectories involving seen objects;
L2, unseen trajectories involving seen objects; and L3, trajectories
involving unseen objects. L1 trajectories are selected heuristically from
the motion-capture datasets based on trajectory and hand-posture similarity
to the training references. For adaptation, we jointly train a single policy
on 20 unseen trajectories, either by fine-tuning the pretrained DexMimic
base policy or by training from scratch.

Fig.~\ref{fig:generalization}(a) reveals a clear scaling trend: increasing
the training set from 20 to 500 trajectories consistently improves
zero-shot success rates across all three generalization levels.
With 500 training trajectories, DexMimic achieves success rates of
68.50\%, 55.89\%, and 51.65\% on L1, L2, and L3, respectively,
up from 8.00\%, 10.63\%, and 2.00\% with 20 trajectories.
These substantial gains show that scaling co-training strengthens the
learned interaction prior, enabling increasingly effective transfer to
unseen trajectories and objects.

Furthermore, the pretrained DexMimic base policy enables rapid adaptation
to unseen trajectories. Figure~\ref{fig:generalization}(b) shows that fine-tuning
reaches 80\% mean training SR across the 20 unseen trajectories
$5.3\times$ faster than training from scratch.
This result highlights the value of pretraining for acquiring new
tracking behaviors efficiently.
We believe that further scaling the training data can yield more capable
generalist policies with stronger generalization and adaptation abilities.

\begin{figure*}[t]
  \begin{minipage}[t]{\columnwidth}
    \vspace{0pt}
  \centering
  \includegraphics[width=\linewidth]{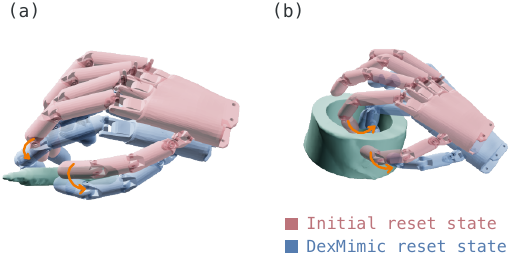}
  \caption{Correction of physically infeasible initial reset states during
  training. Pink and blue show initial reset states and the corresponding
  states updated by DexMimic as training progresses, respectively. Orange
  arrows highlight the fingertip corrections.}
  \label{fig:reset-state-comparison}
  \end{minipage}\hfill
  \begin{minipage}[t]{\columnwidth}
    \vspace{0pt}
  \centering
  \includegraphics[width=\linewidth]{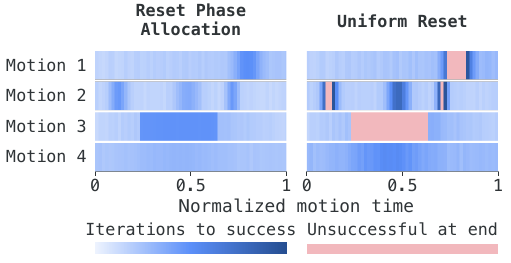}
  \caption{Comparison of reset phase allocation and uniform phase sampling
  on four representative trajectories. Color indicates normalized training
  iterations to success; light red regions remain unsuccessful at the
  end of the same training budget.}
  \label{fig:reset-phase-allocation}
  \end{minipage}
\end{figure*}

\subsection{How Does Value-Guided Reset Allocation Improve Generalist
Training?}

We separately ablate reset state allocation and reset phase allocation
to examine how initialization quality and phase sampling affect generalist
training. RSI initializes from kinematically retargeted reference states
while retaining reset phase allocation; Uniform retains reset state
allocation and samples reference phases uniformly. All three variants
are jointly trained on the same 500 reference trajectories.
Table~\ref{tab:ablation_reset} shows that
DexMimic achieves the highest SR and the lowest object and hand tracking
errors, while reducing training time from 18.60~min to 5.76~min relative to
Uniform, a reduction of 69.0\%.

\begin{table}[t]
  \centering
  \begin{threeparttable}
    \centering
    \footnotesize
    \setlength{\tabcolsep}{2pt}
    \renewcommand{\arraystretch}{1.12}
    \begin{tabular*}{\columnwidth}{@{\extracolsep{\fill}}lcccccccc@{}}
      Method &
      \shortstack{SR $\uparrow$\\(\%)} &
      \shortstack{Time $\downarrow$\\(h)} &
      \shortstack{$E_r \downarrow$\\(rad)} &
      \shortstack{$E_t \downarrow$\\(mm)} &
      \shortstack{$E_j \downarrow$\\(mm)} &
      \shortstack{$E_{ft} \downarrow$\\(mm)} &
      \shortstack{FastFail $\downarrow$\\(\%)} &
      \shortstack{WorstTraj\\SR $\uparrow$} \\
      \midrule
      RSI
        & 66.51
        & 0.150
        & 0.28
        & 5.17
        & 22.64
        & 20.26
        & \cellcolor{weakcell}96.34
        & $21/50$ \\
      Uniform
        & 82.59
        & 0.310
        & 0.18
        & 3.81
        & 20.89
        & 18.39
        & \cellcolor{bestcell}35.65
        & \cellcolor{weakcell}$0/50$ \\
      \midrule
      DexMimic
        & \cellcolor{bestcell}84.73
        & \cellcolor{bestcell}0.096
        & \cellcolor{bestcell}0.17
        & \cellcolor{bestcell}3.55
        & \cellcolor{bestcell}20.50
        & \cellcolor{bestcell}17.38
        & 41.70
        & \cellcolor{bestcell}$38/50$ \\
    \end{tabular*}
    \caption{Ablations validate both reset allocation designs.
    FastFail and WorstTraj SR show that DexMimic addresses two key
    bottlenecks in generalist training: fast-fail resets and learning
    imbalance across trajectories.}
    \label{tab:ablation_reset}
  \end{threeparttable}

\end{table}

\begin{table}[t]
  \centering
  {\fontsize{8}{9.6}\selectfont
  \setlength{\tabcolsep}{4pt}
  \renewcommand{\arraystretch}{1.12}
  \begin{threeparttable}
    \begin{tabular*}{\columnwidth}{@{\extracolsep{\fill}}lrrrr@{}}
      Method & \multicolumn{4}{c}{SR (\%, $\uparrow$) / Time (min, $\downarrow$)} \\
             & 20 & 50 & 200 & 500 \\
      \midrule
      RSI & 74.20 / 30.46 & 72.40 / 15.68
        & 78.46 / 14.32 & 66.51 / 9.00 \\
      Uniform & 76.90 / 24.00 & 67.64 / 17.43
        & 77.79 / 25.89 & 82.59 / 18.60 \\
      \midrule
      DexMimic & \cellcolor{bestcell}85.80 / 20.39
        & \cellcolor{bestcell}75.44 / 10.25
        & \cellcolor{bestcell}88.86 / 8.63
        & \cellcolor{bestcell}84.73 / 5.76 \\
    \end{tabular*}
    \caption{Reset allocation variants trained on 20--500 trajectories.
    Each entry reports SR (\%, $\uparrow$) / average training time
    per trajectory (min, $\downarrow$), using the same timing protocol
    as the main comparison.}
    \label{tab:ablation_scaling}
  \end{threeparttable}}

\end{table}

Compared with RSI, enabling reset state allocation increases SR from
66.51\% to 84.73\% and reduces FastFail from 96.34\% to 41.70\%.
FastFail measures the percentage of all training failures that occur
within 60 frames after reset. Fig.~\ref{fig:reset-state-comparison} illustrates how
Value-Guided Reset Allocation corrects physically infeasible initial
reset states as training progresses. These results show that improving
initialization feasibility mitigates fast-fail resets, allowing the
policy to collect sustained interaction experience.

Compared with Uniform, enabling reset phase allocation increases the
minimum per-trajectory success rate (WorstTraj SR) from 0.00\% to
76.00\%, while improving average SR from 82.59\% to 84.73\%.
The gap in worst-trajectory performance highlights learning imbalance
across trajectories that is obscured by average SR.
Fig.~\ref{fig:reset-phase-allocation} illustrates phase-wise learning
outcomes for four representative trajectories.
These results show that reset phase allocation improves the performance
of the least successful trajectories, addressing learning imbalance under
uniform phase sampling.

We further evaluate both reset allocation mechanisms across co-training
scales of 20, 50, 200, and 500 trajectories. Table~\ref{tab:ablation_scaling}
shows that DexMimic achieves higher success rates and lower average
training time per trajectory than both ablated variants at every
evaluated scale. Reset state allocation yields its largest success-rate
improvement at 500 trajectories, outperforming RSI by 18.22 percentage
points. The efficiency benefit of reset phase allocation grows with
training scale: compared with Uniform, the reduction in average training
time per trajectory increases from 15.0\% at 20 trajectories to 69.0\%
at 500 trajectories. Together, these results show that feasible
initialization supports successful tracking, while adaptive phase
sampling improves training efficiency, enabling generalist tracking to
scale to larger trajectory sets.

\subsection{Real-World Deployment}
\label{sec:realworld-deployment}

As illustrated in Fig.~\ref{fig:realworld_setup}, our real-world platform
consists of a 7-DoF xArm equipped with a 20-DoF Wuji right hand
and a RealSense D435i RGB-D camera. We execute hand--object interaction
trajectories on the physical robot through open-loop replay of successful
simulation rollouts generated by DexMimic. Before execution, we estimate
the initial object pose using FoundationPose++~\cite{yan2025foundationposeplusplus,wen2024foundationpose} and determine a target
wrist pose that matches the hand--object relative pose in the first frame
of the selected rollout. We use PyRoki~\cite{kim2025pyroki} to solve inverse kinematics to align
the arm flange with the initial wrist pose and follow the wrist trajectory
during replay, while the hand executes the corresponding finger joint
motions. We do not enforce strict temporal alignment with the original
human demonstration during replay, as the real robot cannot always match
the speed of human hand motions.

\begin{table}[t]
  \centering
  \fontsize{8}{9.6}\selectfont
  \setlength{\tabcolsep}{4pt}
  \renewcommand{\arraystretch}{1.12}
  \begin{threeparttable}
  \begin{tabular*}{\columnwidth}{@{\extracolsep{\fill}}lrrrr@{}}
    Hand & Wuji & Inspire & Shadow & Allegro \\
    \midrule
    SR $\uparrow$ & 872/1000 & 861/1000
      & 832/1000 & 759/1000 \\
  \end{tabular*}
  \caption{DexMimic across dexterous hands.}
  \label{tab:cross-hand}
  \end{threeparttable}
  \par\vspace{8pt}
  \begin{minipage}{\columnwidth}
    \includegraphics[width=\linewidth]{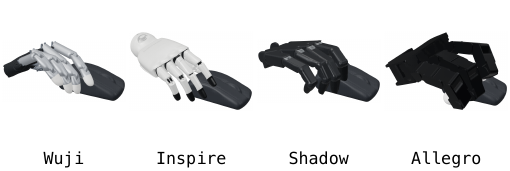}
    \captionof{figure}{Using mouse with Wuji, Inspire, Shadow, and Allegro.}
    \label{fig:cross-hand}
  \end{minipage}
\end{table}

\subsection{Performance across Dexterous Hands}
\label{sec:cross-hand}

We further evaluate DexMimic on four dexterous hands, Wuji, Inspire,
Shadow, and Allegro, as illustrated in Fig.~\ref{fig:cross-hand}. We train one policy
for each hand. Each policy is trained on the same set of 20 dexterous
manipulation tasks. Evaluation follows the protocol in
Sec.~\ref{sec:experimental-settings}. Table~\ref{tab:cross-hand} reports
the overall success rate (SR) for each hand over 1,000 evaluation rollouts.

\section{Conclusion}
We believe that learning dexterous manipulation from large-scale human hand--object interaction datasets offers a promising pathway toward human-level robot dexterity. DexMimic addresses the reset allocation bottleneck in scaling reference-conditioned tracking to a generalist policy through Value-Guided Reset Allocation. Experiments demonstrate improved tracking success and training efficiency, faster adaptation, zero-shot generalization potential, and transfer of learned behaviors to physical robots.

Although DexMimic shows promise in learning diverse dexterous skills from human demonstrations, it remains constrained by the scale of available HOI datasets. Future work will explore tracking methods that are more robust to noise in reconstructed hand--object interactions~\cite{hasson2021homan,fan2023hold,liu2025easyhoi,wang2025magichoi}, enabling learning from more abundant in-the-wild videos.

\bibliographystyle{IEEEtran}
\bibliography{references}

\end{document}


\raggedbottom
\ifdefined\arxivsupplement
\setcounter{page}{\arxivfirstpage}
\pagestyle{plain}
\twocolumn[{{\centering\fontsize{20}{24}\selectfont\bfseries Supplemental Material\par}\vspace{8pt}\suppoverview\vspace{8pt}}]
\else
\maketitle
\fi

\section{Policy and Environment Details}

\noindent\textbf{Notation.}
Let $\tau_i=\{\hat{x}_{i,k}\}_{k=0}^{T_i}$, $s_t$, $o_t$, and $a_t$
denote the reference trajectory, simulator state, observation, and
residual action. Superscripts $w$ and $o$ index the wrist rigid body
and manipulated object, respectively. For $b\in\{w,o\}$, let
$(\boldsymbol p_t^b,\boldsymbol Q_t^b,
\boldsymbol v_t^b,\boldsymbol\omega_t^b)$ denote its simulated position,
orientation quaternion, linear velocity, and angular velocity;
hatted quantities are reference values from $\hat{x}_{i,k}$.
For $x\in\{p,v,\omega\}$, define target-minus-simulation errors as
\begin{equation}
\boldsymbol e^{b,x}_{i,t,k}=\hat{\boldsymbol x}_{i,k}^b-\boldsymbol x_t^b.
\label{eq:supp-tracking-errors}
\end{equation}
The orientation difference is
\begin{equation}
\boldsymbol e^{b,Q}_{i,t,k}=\boldsymbol Q_t^b\otimes
(\hat{\boldsymbol Q}_{i,k}^b)^{-1}.
\label{eq:supp-orientation-error}
\end{equation}
Here $\otimes$ denotes quaternion multiplication; $\boldsymbol q_t$ and
$\dot{\boldsymbol q}_t$ are finger joint positions and velocities.
Index $j$ pairs a reference hand point with its simulated rigid body.

\subsection{Observation Space}

The observation contains histories of hand proprioception and hand--object
interaction features, together with reference targets. History and future offsets in control
frames are
\begin{equation}
\mathcal H=\{0,-1,-2,-4,-8\},\qquad
\mathcal F=\{0,1,2,4,8\}.
\label{eq:supp-offsets}
\end{equation}
The proprioceptive vector at frame $t$ is
\begin{equation}
\boldsymbol h_t=
[\boldsymbol q_t,\cos\boldsymbol q_t,\sin\boldsymbol q_t,
\boldsymbol p_t^w,\boldsymbol Q_t^w,
\boldsymbol v_t^w,\boldsymbol\omega_t^w].
\label{eq:supp-proprioception}
\end{equation}
The wrist position is retained in this block, and the object position
relative to the wrist is included below. Missing history frames repeat
the reset state for both feature blocks.
Let $\boldsymbol f_t^j$ denote fingertip contact force,
$\boldsymbol p_t^{\mathrm{CoM}}$ the object center of mass, and $w_o$
its weight. The interaction features at frame $t$ are
\begin{equation}
\boldsymbol c_t=[\dot{\boldsymbol q}_t,
\boldsymbol c_t^{\mathrm{obj}},\boldsymbol c_t^{\mathrm{tip}}].
\label{eq:supp-interaction}
\end{equation}
\begin{equation}
\boldsymbol c_t^{\mathrm{obj}}=[\boldsymbol p_t^o-\boldsymbol p_t^w,
\boldsymbol p_t^o,\boldsymbol Q_t^o,\boldsymbol v_t^o,\boldsymbol\omega_t^o,
\boldsymbol p_t^{\mathrm{CoM}}-\boldsymbol p_t^w,w_o].
\label{eq:supp-object-state}
\end{equation}
\begin{equation}
\boldsymbol c_t^{\mathrm{tip}}=
\{\boldsymbol f_t^j,\|\boldsymbol f_t^j\|_2\}_{j=1}^{5}.
\label{eq:supp-contact-state}
\end{equation}

For $k=\min(t+\delta,T_i)$ and $\delta\in\mathcal F$, the wrist and
object reference blocks are
\begin{equation}
\boldsymbol r^b_{i,t,k}=
[\boldsymbol r^{b,\mathrm{lin}}_{i,t,k},
\boldsymbol r^{b,\mathrm{rot}}_{i,t,k}].
\label{eq:supp-body-reference}
\end{equation}
\begin{equation}
\boldsymbol r^{b,\mathrm{lin}}_{i,t,k}=
[\boldsymbol e^{b,p}_{i,t,k},\hat{\boldsymbol v}_{i,k}^b,
\boldsymbol e^{b,v}_{i,t,k}].
\label{eq:supp-linear-reference}
\end{equation}
\begin{equation}
\boldsymbol r^{b,\mathrm{rot}}_{i,t,k}=
[\hat{\boldsymbol Q}_{i,k}^b,\boldsymbol e^{b,Q}_{i,t,k},
\hat{\boldsymbol\omega}_{i,k}^b,\boldsymbol e^{b,\omega}_{i,t,k}].
\label{eq:supp-rotation-reference}
\end{equation}
Reference hand points contribute position errors, target velocities,
and velocity errors:
\begin{equation}
\boldsymbol r^{\mathrm{link}}_{i,t,k}=
\{\hat{\boldsymbol p}_{i,k}^j-\boldsymbol p_t^j,
\hat{\boldsymbol v}_{i,k}^j,
\hat{\boldsymbol v}_{i,k}^j-\boldsymbol v_t^j\}_{j=1}^{J}.
\label{eq:supp-link-reference}
\end{equation}
Let $\hat{\boldsymbol d}^{\mathrm{tip}}_{i,k}$ contain reference
fingertip--object distances, and define
$\boldsymbol h_t^{\mathcal H}=
[\boldsymbol h_{t+\delta}]_{\delta\in\mathcal H}$,
$\boldsymbol c_t^{\mathcal H}=
[\boldsymbol c_{t+\delta}]_{\delta\in\mathcal H}$, and
$\boldsymbol r_{i,t}^{\mathcal F}=
[\boldsymbol r_{i,t,\min(t+\delta,T_i)}]_{\delta\in\mathcal F}$ using
\begin{equation}
\boldsymbol r_{i,t,k}=
[\boldsymbol r^w_{i,t,k},\boldsymbol r^o_{i,t,k},
\boldsymbol r^{\mathrm{link}}_{i,t,k},
\hat{\boldsymbol d}^{\mathrm{tip}}_{i,k}].
\label{eq:supp-reference-block}
\end{equation}
The full observation is
\begin{equation}
o_t=[\boldsymbol h_t^{\mathcal H},\boldsymbol c_t^{\mathcal H},
\boldsymbol r_{i,t}^{\mathcal F},\boldsymbol d_t^o,
\boldsymbol b_i^o,\boldsymbol a_t^{\mathrm{base}},\ell_t].
\label{eq:supp-observation}
\end{equation}
Here $\boldsymbol d_t^o$ contains object-to-hand rigid-body distances,
$\boldsymbol b_i^o$ is the 128-dimensional object geometry encoding,
$\boldsymbol a_t^{\mathrm{base}}$ is the kinematic base action, and
$\ell_t$ gives the remaining reference steps. Bimanual observations
concatenate hand-specific blocks and share one $\ell_t$; actor and critic
receive the same observation.

\subsection{Action Space}

For $n$ actuated finger joints, the action comprises normalized wrist
translation, wrist rotation, and finger target residuals:
\begin{equation}
\boldsymbol a_t=[\boldsymbol a_t^p,\boldsymbol a_t^R,
\boldsymbol a_t^q]\in[-1,1]^{6+n},\qquad
\boldsymbol\delta_t=2\boldsymbol a_t.
\label{eq:supp-action}
\end{equation}

The kinematic base action comes from the retargeted hand reference.
Its wrist position error $\boldsymbol e^{w,p}_{i,t,t}$ is clipped to
$0.10\,\mathrm{m}$; the axis--angle error of
$\hat{\boldsymbol Q}_{i,t}^w\otimes(\boldsymbol Q_t^w)^{-1}$ is clipped
to $1\,\mathrm{rad}$. The observation uses the reverse quaternion
difference in Eq.~\eqref{eq:supp-orientation-error}.
The finger base $\boldsymbol b_t^q$ is the normalized retargeted target.

In wrist PID control, residuals augment the base force and torque:
\begin{equation}
\widetilde{\boldsymbol F}_t=
\mathrm{PID}_p(\bar{\boldsymbol e}_t^p)
+500\Delta t\,s_p\boldsymbol\delta_t^p.
\label{eq:supp-wrist-force}
\end{equation}
\begin{equation}
\widetilde{\boldsymbol M}_t=
\mathrm{PID}_R(\bar{\boldsymbol e}_t^R)
+200\Delta t\,s_R\boldsymbol\delta_t^R.
\label{eq:supp-wrist-torque}
\end{equation}
Here $\bar{\boldsymbol e}_t^p$ and $\bar{\boldsymbol e}_t^R$ denote the
clipped errors, $\Delta t=1/60\,\mathrm{s}$, $s_p=1$, and $s_R=0.1$;
PID gains depend on the hand model. Finger residuals are added in
normalized joint coordinates and mapped to joint limits:
\begin{equation}
\boldsymbol u_t^q=
\operatorname{clip}(\boldsymbol b_t^q+\boldsymbol\delta_t^q,-1,1).
\label{eq:supp-finger-residual}
\end{equation}
\begin{equation}
\boldsymbol q_t^{\mathrm{raw}}=\boldsymbol q_{\min}+
\tfrac12(\boldsymbol u_t^q+1)\odot
(\boldsymbol q_{\max}-\boldsymbol q_{\min}).
\label{eq:supp-finger-target}
\end{equation}
Wrist force, torque, and finger targets use temporal smoothing with
$\alpha=0.4$:
\begin{equation}
\boldsymbol y_t=\alpha\widetilde{\boldsymbol y}_t
+(1-\alpha)\boldsymbol y_{t-1},\quad
\boldsymbol y\in\{\boldsymbol F,\boldsymbol M,\boldsymbol q^{\mathrm{cmd}}\}.
\label{eq:supp-action-smoothing}
\end{equation}
For finger control, $\widetilde{\boldsymbol q}^{\mathrm{cmd}}_t=
\boldsymbol q_t^{\mathrm{raw}}$; the smoothed target is clipped to joint
limits. Wrist force $\boldsymbol F_t$ (N) and torque
$\boldsymbol M_t$ (N$\cdot$m) act on the wrist rigid body; finger targets
are in radians. Object motion follows contact dynamics. Bimanual actions
concatenate both hands' residuals.

\begin{table}[!t]
\centering
\footnotesize
\renewcommand{\arraystretch}{1.06}
\begin{tabular*}{\columnwidth}{@{\extracolsep{\fill}}llr@{}}
Term (unit) & Component reward & Weight \\
\midrule
\multicolumn{3}{@{}l}{{\small\bfseries Hand pose tracking}} \\
Wrist position (m) & $\exp(-40E_{w,p})$ & $0.10$ \\
Wrist orientation (rad) & $\exp(-E_{w,R})$ & $0.60$ \\
Thumb (m) & $\exp(-100E_T)$ & $0.90$ \\
Index (m) & $\exp(-90E_I)$ & $0.80$ \\
Middle (m) & $\exp(-80E_M)$ & $0.75$ \\
Ring (m) & $\exp(-60E_R)$ & $0.60$ \\
Little (m) & $\exp(-60E_P)$ & $0.60$ \\
Link group $L_1$ (m) & $\exp(-50E_{L_1})$ & $0.50$ \\
Link group $L_2$ (m) & $\exp(-40E_{L_2})$ & $0.30$ \\
\addlinespace[2pt]
\multicolumn{3}{@{}l}{{\small\bfseries Object pose tracking}} \\
Object position (m) & $\exp(-80E_{o,p})$ & $5.00$ \\
Object orientation (rad) & $\exp(-6E_{o,R})$ & $1.00$ \\
\addlinespace[2pt]
\multicolumn{3}{@{}l}{{\small\bfseries Velocity matching}} \\
Wrist linear (m/s) & $\exp(-E_{w,v})$ & $0.10$ \\
Wrist angular (rad/s) & $\exp(-E_{w,\omega})$ & $0.05$ \\
Hand links (m/s) & $\exp(-E_{\mathrm{link},v})$ & $0.10$ \\
Object linear (m/s) & $\exp(-E_{o,v})$ & $0.10$ \\
Object angular (rad/s) & $\exp(-E_{o,\omega})$ & $0.10$ \\
\addlinespace[2pt]
\multicolumn{3}{@{}l}{{\small\bfseries Reference-guided contact}} \\
Fingertip contact (N) & $\exp[-(F_t^{\mathrm{tip}}/\mathrm N+10^{-5})^{-1}]$ & $1.00$ \\
\addlinespace[2pt]
\multicolumn{3}{@{}l}{{\small\bfseries Mechanical-power regularization}} \\
Joint power (W) & $\exp(-10P_t^q)$ & $0.50$ \\
Wrist power (W) & $\exp(-2P_t^w)$ & $0.50$ \\
\end{tabular*}
\caption{Single-hand reward components. Units refer to the
underlying errors or physical quantities.}
\label{tab:supp-reward}
\end{table}

\subsection{Reward Function}

The task-agnostic reward comprises hand pose, object pose, velocity,
reference-guided contact, and mechanical-power terms. For the set
$\mathcal G$ of these groups,
\begin{equation}
r_t=\sum_{g\in\mathcal G}\lambda_g r_t^g.
\label{eq:supp-reward-groups}
\end{equation}
The hand group includes wrist and finger tracking.
Table~\ref{tab:supp-reward} specifies each component's weight;
$\lambda_g=1$ for all five groups.

At the current reference frame, pose errors are
\begin{equation}
E_{b,p}=\|\boldsymbol e^{b,p}_{i,t,t}\|_2,\qquad
E_{b,R}=2\arccos|\langle\boldsymbol Q_t^b,
\hat{\boldsymbol Q}_{i,t}^b\rangle|.
\label{eq:supp-pose-errors}
\end{equation}
For each hand-link group $G$,
\begin{equation}
E_G=|\mathcal I_G|^{-1}\sum_{j\in\mathcal I_G}
\|\hat{\boldsymbol p}_{i,t}^j-\boldsymbol p_t^j\|_2.
\label{eq:supp-hand-link-error}
\end{equation}
Velocity errors average absolute component differences:
\begin{equation}
E_{b,v}=\tfrac13\|\boldsymbol e^{b,v}_{i,t,t}\|_1,\qquad
E_{b,\omega}=\tfrac13\|\boldsymbol e^{b,\omega}_{i,t,t}\|_1.
\label{eq:supp-velocity-errors}
\end{equation}
For tracked hand links,
\begin{equation}
E_{\mathrm{link},v}=(3J)^{-1}\sum_{j=1}^{J}
\|\hat{\boldsymbol v}_{i,t}^j-\boldsymbol v_t^j\|_1.
\label{eq:supp-link-velocity-error}
\end{equation}

Reference fingertip--object distance gates the simulated fingertip force.
For fingertip $j$, set
$u_t^j=\operatorname{clip}((0.03-\hat d^{\mathrm{tip},j}_{i,t})/0.01,0,1)$,
with distances in metres, and let
\begin{equation}
F_t^{\mathrm{tip}}=\sum_{j=1}^{5}u_t^j\|\boldsymbol f_t^j\|_2.
\label{eq:supp-contact-force}
\end{equation}
Mechanical-power terms use
\begin{equation}
P_t^q=\sum_{j=1}^{n}|\tau_{t,j}^q\dot q_{t,j}|.
\label{eq:supp-joint-power}
\end{equation}
\begin{equation}
P_t^w=|\boldsymbol F_t^\top\boldsymbol v_t^w|
+|\boldsymbol M_t^\top\boldsymbol\omega_t^w|.
\label{eq:supp-wrist-power}
\end{equation}
Component rewards and weights are dimensionless; exponential scales
assume SI units. Bimanual rewards sum per-hand terms, including an
object term for each hand.

\begin{figure}[!t]
\centering
\includegraphics[width=\columnwidth]{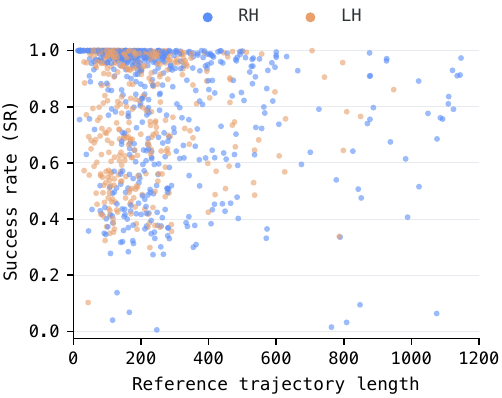}
\caption{Per-trajectory success rates versus reference trajectory length. Colors distinguish right-hand (RH) and left-hand (LH) trajectories.}
\label{fig:supp-trajectory-sr}

\end{figure}

\begin{table}[!t]
\centering
\footnotesize
\begin{tabular*}{\columnwidth}{@{\extracolsep{\fill}}ll@{}}
\toprule
Parameter & Value \\
\midrule
PPO rollout / mini-epochs / minibatch & $32$ / $5$ / $1024$ \\
Discount $\gamma$ / GAE $\lambda_{\mathrm{GAE}}$ & $0.99$ / $0.95$ \\
Adam learning rate / PPO clip & $5\!\times\!10^{-4}$ / $0.2$ \\
Reset slots $K$ / anchor probability $\alpha$ & $3$ / $0.1$ \\
Survival threshold $c_{\min}$ / score decay $\delta$ & $0.1$ / $10^{-5}$ \\
Bin width / lookback $n$ / factor $\lambda$ & $15$ / $5$ / $0.8$ \\
\bottomrule
\end{tabular*}
\caption{Training and reset allocation hyperparameters.}
\label{tab:supp-hyperparameters}
\end{table}

\section{Training and Reset Allocation Details}

\subsection{PPO and Network Configuration}

Each observation block is normalized using running statistics before
concatenation. A feature MLP with three $512$-unit Swish layers and a
$256$-dimensional projection feeds a shared ELU trunk of widths
$256$, $512$, $128$, and $64$. Separate heads output the mean of a
diagonal Gaussian policy and the scalar critic $V_\phi(o_t)$; the
Gaussian has trainable, observation-independent log standard deviations.
Policy samples are bounded to $[-1,1]$ before residual action synthesis.
Clipped PPO jointly updates the shared features and both heads using
generalized advantage estimation. Value targets are normalized during
optimization, while reset allocation uses critic estimates on the
original reward scale.

\subsection{Reset Buffer Update}

For each reference phase $(i,t)$, the reset buffer $\mathcal B_{i,t}$
holds $K$ scored states, initialized from the phase-aligned reference
state $s_{i,t}^{\mathrm{ref}}$ and its critic score. The fixed reference
anchor is sampled through a separate $\alpha$ branch. Each state stores
hand and object root configurations and velocities, finger joint
positions and velocities, and controller targets. Before action
execution at relative step $u$ of episode $e$,
the physical state is paired with its critic estimate. At episode end,
the pair is assigned to the corresponding reference frame:
\begin{equation}
t=t_e^{\mathrm{start}}+u,\qquad v_{e,u}=V_\phi(o_{e,u}).
\label{eq:supp-candidate-alignment}
\end{equation}
For candidates assigned to the same phase, scores are processed in
descending order. Here $c(s)$ is the fraction of remaining reference
steps tracked from the candidate before episode termination. A candidate
must satisfy $c(s)>c_{\min}$ and strictly exceed the current lowest
buffer score to replace that entry; the lowest score is recomputed after each
replacement. After a nonempty episode-finalization batch, candidate
scores decay once as $v\leftarrow(1-\delta)v$. The reference anchor
remains fixed and is used if all candidate sampling weights vanish.

\subsection{Reset Phase Allocation}

Each reference trajectory is partitioned into temporal bins of $15$
control frames; its final bin may be shorter. The bins are indexed
globally across trajectories. For each rollout step with $\ell_t>0$,
the critic estimate on the original reward scale is normalized by the
remaining discounted horizon:
\begin{equation}
\nu_t=\frac{V_\phi(o_t)}{\sum_{h=0}^{\ell_t-1}\gamma^h}.
\label{eq:supp-value-rate}
\end{equation}
For bins receiving samples, let $\mathcal D_b$ be the multiset of
rollout steps assigned to bin $b$, including repeated visits. Their mean is
$\bar\nu_b=|\mathcal D_b|^{-1}
\sum_{t\in\mathcal D_b}\nu_t$. The difficulty
$f_b=\nu_{\max}-\bar\nu_b$ uses the theoretical maximum normalized
value implied by the reward. It propagates to preceding bins within
the same trajectory:
\begin{equation}
\tilde f_b=\sum_{d=0}^{n-1}\lambda^d f_{b+d},
\label{eq:supp-phase-lookback}
\end{equation}
with terms crossing the trajectory boundary omitted. The propagated
score updates the phase priority by exponential moving average:
\begin{equation}
\bar f_b\leftarrow(1-\beta)\bar f_b+\beta\tilde f_b.
\label{eq:supp-phase-ema}
\end{equation}
Sampling normalizes $\bar f_b$ over all bins, then selects uniformly
among the $n_b$ valid frames of the chosen bin. Thus, for a frame
$(i,t)$ in bin $b(i,t)$,
\begin{equation}
p_{\mathrm{phase}}(i,t)=
\frac{\bar f_{b(i,t)}}{n_{b(i,t)}\sum_{j=1}^{B}\bar f_j}.
\label{eq:supp-phase-probability}
\end{equation}

\subsection{Hyperparameters}

Table~\ref{tab:supp-hyperparameters} summarizes the principal training and reset parameters.

\flushcolsend
\raggedend
\twocolumn[{
\begin{minipage}{\textwidth}
\centering
\includegraphics[width=\textwidth]{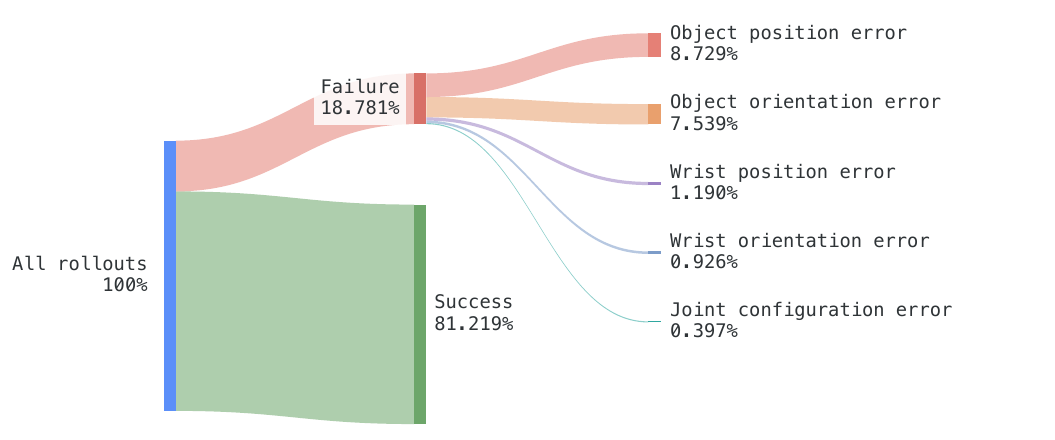}
\captionof{figure}{Evaluation outcomes and estimated failure-category proportions.
Failure labels denote tracking errors exceeding their respective thresholds.
All percentages are relative to the 1,017 evaluation rollouts.
Failure-category proportions are estimated by proportionally allocating
unclassified failures across the five categories.}
\label{fig:supp-failure}
\par\vspace{6pt}
\includegraphics[width=\textwidth]{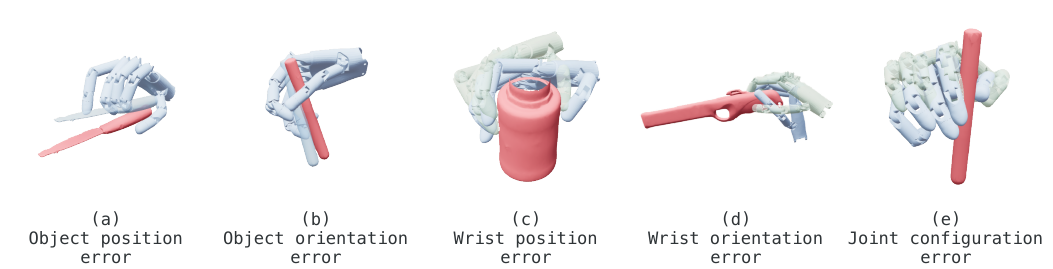}
\captionof{figure}{Representative examples of the five failure categories. Translucent hands and objects indicate the reference.}
\label{fig:supp-failure-cases}

\end{minipage}
\vspace{10pt}
}]

\section{Additional Results}

\subsection{Per-Trajectory Success Rates}

We evaluate DexMimic on 1,302 trajectories from OakInk and HOT3D.
Figure~\ref{fig:supp-trajectory-sr} plots success rate against reference
trajectory length, with colors distinguishing right-hand (RH) and
left-hand (LH) trajectories. Figure~\ref{fig:supp-rollouts} shows representative
states from 28 recorded DexMimic trajectories, including 23 single-hand
and five bimanual examples.

\newpage
\subsection{Failure Case Analysis}

Across 1,017 evaluation rollouts on a subset of OakInk trajectories,
DexMimic succeeds in 826 cases (81.219\%). Rollouts terminate early when tracking errors in object
position, object orientation, wrist position, wrist orientation, or
finger joint configuration exceed their respective thresholds.
The estimated breakdown in Fig.~\ref{fig:supp-failure} is dominated
by object tracking failures, indicating that maintaining the reference
object position and orientation remains the primary challenge in
these evaluations. Figure~\ref{fig:supp-failure-cases} shows
representative cases for the five categories.